# Hybrid Generative/Discriminative Learning for Automatic Image Annotation

**Shuang Hong Yang**
College of Computing
Georgia Institute of Technology
shy@gatech.edu

**Jiang Bian**
College of Computing
Georgia Institute of Technology
jbian@cc.gatech.edu

**Hongyuan Zha**
College of Computing
Georgia Institute of Technology
zha@cc.gatech.edu

## Abstract

Automatic image annotation (AIA) raises tremendous challenges to machine learning as it requires modeling of data that are both ambiguous in input and output, e.g., images containing multiple objects and labeled with multiple semantic tags. Even more challenging is that the number of candidate tags is usually huge (as large as the vocabulary size) yet each image is only related to a few of them. This paper presents a hybrid generative-discriminative classifier to simultaneously address the extreme data-ambiguity and overfitting-vulnerability issues in tasks such as AIA. Particularly: (1) an Exponential-Multinomial Mixture (EMM) model is established to capture both the input and output ambiguity and in the meanwhile to encourage prediction sparsity; and (2) the prediction ability of the EMM model is explicitly maximized through discriminative learning that integrates variational inference of graphical models and the pairwise formulation of ordinal regression. Experiments show that our approach achieves both superior annotation performance and better tag scalability.

## 1 Introduction

As the exponential growth of internet photographs (e.g. Flickr) and videos (e.g. Youtube), *automatic image annotation* (AIA) is increasingly important for indexing, managing and retrieving multimedia data. AIA usually includes two types of tasks: (1) *image annotation* assigns descriptive metadata (e.g. caption) to a given (entire) image; and (2) *region annotation* annotates each object (e.g., image region) within a given image with appropriate textual tags.

AIA raises tremendous challenges to machine learning algorithms. Firstly, because each image usually contains multiple objects in a consistent scene, to recognize each object thus requires modeling of input-ambiguous systems, i.e., each input example (e.g., image) contains a set of correlated instances (e.g., objects), for which each individual output (e.g., tag prediction) is of interest. This triggers the nontrivial application of multi-instance learning (Dietterich et al, 1998; Carneiro et al, 2007). In addition, due to the huge image volumes and extremely diverse scene topics, it is prohibitive to employ skilled experts to manually label a set of training data that are sufficiently large to support statistical learning. Instead, we only have access to partially labeled images, e.g., photos with captions specifying only names of persons in each photo but not which name goes with which face. Learning from such ambiguously labeled multi-instance data opens a new frontier for machine learning, i.e., modeling systems/data that are both ambiguous in input and output, motivating the recently emerged framework of *multi-label multi-instance classification* (Cour et al, 2009; Yang et al, 2010; Zhou and Zhang, 2007).

Perhaps even more challenging in AIA is that, in a real world AIA task, the number of candidate tags is usually huge (as large as the size of a natural language vocabulary) yet each image is only related to a few of them. Indeed, because the scene topics of internet images/videos are extremely diverse, any noun-phrase can be a valid tag. This makes naive classifier extremely vulnerable to overfitting because (1) that a few relevant tags could easily be eclipsed in the crowd of a huge number of irrelevant ones; and (2) that the data could be extremely scarce for training the classifier corresponding to each individual tag. For example, considering a setting consists of 1000 evenly distributed tags, even with 100K labeled training images, we only have 100 examples for each tag, and 0.1% expected accuracy for a baseline random guessing classifier. To reduce the risk of overfitting, it is desirable to pursue classifiers that are scalable to class (tag) size. However, little progress has been made so far on this topic.

In this paper, we abstract AIA as a generic learning task called *many-class multi-label multi-instance classification* ($M^3C$), which aims at learning decision rules from data that (1) are ambiguous in both input and output and (2) involve massive classes. Typically, in an $M^3C$ data set, each example (e.g. image) consists of multiple instances

(e.g. objects) and is associated with several out of a huge number of classes (e.g. tags). In this paper, we present a hybrid generative-discriminative classifier to address both the extreme data ambiguity and overfitting vulnerability (due to massive classes) issues involved in $M^3C$. Particularly, an *Exponential-Multinomial Mixture* (EMM) model is established, which is able to capture both the input and output ambiguity of $M^3C$ data, and in the meanwhile to encourage prediction sparsity so as to automatically rule out irrelevant classes (e.g., tags). Furthermore, to maximize discriminant ability and to handle the *small-sample problem* incurred by the massive classes as well, we derive inference and learning algorithms based on the principle of margin maximization (Sha and Saul, 2007; Zhu et al, 2009), leading to a learning formulation integrating the *variational inference* of graphical model (Jaakkola and Jordan, 2000) and the pariwise preference formulation of *ordinal regression* (Herbrich et al, 1999), which is further efficiently solved by convex optimization. Compared with traditional MLE learning, the discriminative learning formulation not only explicitly optimizes the predication (classification) performance, but also enjoys additional advantages such as stronger supervision and perfect data balance.

Our AIA system builds the multi-instance multi-label corpora by (1) segmenting each image into instances (i.e. regions) and (2) using words in the caption as labels. Each region is represented using the standard bag-of-discrete-feature framework (Nowak et al, 2006). In this way, an example (i.e., image) is analogous to a document, an instance (i.e., region) to a paragraph, and a feature to a word. Experiments on image and caption collections from `Alipr` and `LabelMe` show that our algorithm achieves not only superior annotation accuracy but also better tag-scalability, i.e., performance robustness against (1) label size increasing, or (2) training size decreasing.

## 2 Related Works

Automatic image annotation is usually addressed using machine translation approaches (Blei and Jordan, 2003; Duygulu et al, 2002; Li and Wang, 2008; Wang et al, 2009), in which image-caption pairs are viewed as bi-lingual texts and machine translation techniques are applied to align (translate) the textual vocabulary (e.g., tags) with the visual vocabulary (e.g., blobs clustered from image regions). In this paper, instead of learning the loose correspondences between tags and visual words, we attempt to learn a classifier that is able to directly discriminate the semantic tags based on the visual content/context of the image regions.

There are two critical challenges in AIA, one is the data ambiguity (in both input and output), the other is the overfitting vulnerability due to massive tags. Substantial efforts have been made to address the former issue. Traditional multi-class image classification assigns a single tag to an entire image and thus captures neither the input nor the output ambiguity. Multi-label classification (Carneiro et al, 2007), annotating each image with one or more tags, only captures output ambiguity and is applicable only to task (1) (i.e., image annotation); whereas multi-instance classification (Viola et al, 2006; Yang et al, 2005), which annotates each image region with a single tag, only captures input ambiguity and thus cannot annotate multi-tag images. Quite recently, several approaches (Cour et al, 2009; Zha et al, 2009; Yang et al, 2010) were developed for multi-label multi-instance classification, which naturally address both input and output ambiguity. Although all these approaches were observed to achieve superior performances for ambiguity modeling (though not directly applied to AIA), none of them addresses the massive-tag issue, which is a fundamental challenge of real-world tasks. As a result, they are only applicable to task with moderate number of classes. In this paper, we attempt to simultaneously address both the data ambiguity and massive-tag issues. Our model builds on the Dirichlet-Bernoulli Alignment (DBA) model (Yang et al, 2010) to capture data ambiguity. DBA is a Dirichlet-Multinomial mixture model, similar to the sLDA (Blei et al, 2003) topic model but with supervised topics (topics explicitly aligned to class labels). A limitation of DBA is that the classification prediction turns to be smoothed over all the classes, making its performance deteriorate exponentially as the number of classes increases. In this paper, we take two measures to improve tag-scalability. First, instead of using Dirichlet prior to smooth prediction among tags (e.g., sLDA and DBA), we adopt Exponential prior to encourage prediction sparsity. This significantly restricts the number of active tags for a given image, ensuring the classifier effectively identify the most relevant tags while automatically ruling out irrelevant ones. Secondly, unlike traditional MLE learning, our model is trained discriminatively to explicitly maximize prediction accuracy.

We also noticed a surge of needs for sparse topic models in machine learning community. We believe our model is a good start. In its simplified version, our model seamlessly integrates two very important learning tools: *topic modeling* and *sparse coding*, and enjoys the advantages of both: the outstanding interpretability of topic models in discovering topics that are intuitively comprehensible, and the extraordinary performance of sparse-coding in learning predictive topics. Indeed, sparsity turns to be an highly-preferable property (or even requirement) for learning algorithms. In some cases, the objective (likelihood) is not convex or the solution space is flat in some areas, nonsparse learning algorithms usually lead to local optima or under-learned models. With sparsity prior/regularization/constraint, the solutions become more distinguishable (i.e., only solutions in or close to the surfaces of the hyper-polyhedron are preferred) and hence lead to better generalization ability. From the information-theoretic viewpoint, being sparser might also imply smaller description length.

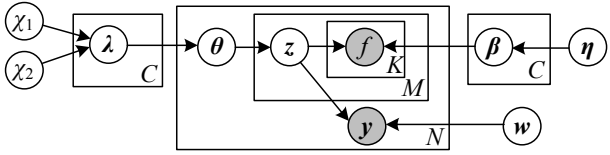

Figure 1: The Exponential-Multinomial Mixture (EMM) model.

## 3 Hybrid Generative/Discriminative Learning for M$^3$C

### 3.1 Formulation

In traditional multiclass classification, we are given a set of labeled examples $\{\mathbf{x}_n, y_n\}_{n=1}^N$, where $\mathbf{x}_n$ is a point in a vector space $\mathcal{X} \subset \mathbb{R}^D$, while $y_n$ is a point in a moderate-sized set $\mathcal{Y} = \{1, 2, \ldots, C\}$ ($C > 2$). The goal is to find a decision rule $y = \phi(x) : \mathcal{X} \to \mathcal{Y}$.

In a *Many-Class Multi-Label Multi-Instance Classification* (M$^3$C) task such as AIA, the number of classes are extremely large $C \gg 2$, and each example as well as its label are neither points but both sets, that is: given a labeled training set $\mathcal{D} = \{\mathsf{X}_n, \mathsf{Y}_n\}_{n=1}^N$, each example (e.g., image) $\mathsf{X}_n$ contains a set of instances (e.g., objects within the image) $\mathbf{x}_{mn} \in \mathcal{X}$ ($m = 1, 2, \ldots, M_n$), and its label (e.g., caption of the image) $\mathsf{Y}_n \subset \mathcal{Y}$ consists of a set of class labels (e.g., tag words in the caption). The goal of M$^3$C is to find a decision rule $\mathsf{Y} = \varphi(\mathsf{X}) : 2^\mathcal{X} \to 2^\mathcal{Y}$, where $2^\mathcal{A}$ denotes the power set of a set $\mathcal{A}$. We make the the following assumptions:

- **Exchangeability**: A data set is a bag of examples (e.g. images), and each example is a bag of instances (e.g. objects within an image).
- **Parsimoniousness**: Each example is related to a very limited number of classes, i.e., $\forall n, |\mathsf{Y}_n| \ll C$.
- **Distinguishablity**: Each instance (region) within an example belongs to a single class.

### 3.2 Exponential-Multinomial Mixture

In this section, we develop a Bayesian hierarchical model to model the data ambiguity in M$^3$C. The model is built on recent advances in probabilistic topic models (Blei et al, 2003; Blei and Mcauliffe, 2008). By modeling a document as an admixture of latent topics, topic models allow each document to be associated with multiple topics with different proportions, and thus provide a principled way to capture the ambiguity in the data. However, the topics discovered by topic model are essentially multinomial distributions over words. Although for textual data, it is possible to interpret the discovered topics based on human expertise (Blei et al, 2003), for image data, multinomial distributions of visual features are generally incomprehensible. Here, we use natural language tags as explicit topics, which not only allows us to learn classifiers (rather than dimensionality reducers) from ambiguously labeled data, but also enables discriminative learning. Also, by using exponential rather than Dirichlet prior, our model seamlessly integrates two important learning tools, i.e., *topic modeling* (Blei et al, 2003) and *sparse coding* (Lee et al, 2009).

We develop a Bayesian hierarchical model for M$^3$C based on the aforementioned assumptions. The Exponential-Multinomial Mixture (EMM) model (Figure 1) assume that each example $\mathsf{X}$ in $\mathcal{D}$ is generated by the following process:

1. Sample $\boldsymbol{\theta} \sim \text{Exp}(\boldsymbol{\lambda})$
2. For each of the $M$ instances in $\mathsf{X}$:
   a) Choose a label $\mathbf{z} \sim \text{Mult}(\tilde{\boldsymbol{\theta}})$;
   b) Generate an instance $\mathbf{x} \sim p(\mathbf{x}|\mathbf{z}, B)$;
3. Generate the example-level label $\mathbf{y} \sim p(\mathbf{y}|\mathbf{z}_{1:M}, \mathbf{w})$.

In the model, an exponential distribution $p(\boldsymbol{\theta}|\boldsymbol{\lambda})$ is used to model the property of label parsimoniousness, where $\boldsymbol{\lambda} = [\lambda_1, \ldots, \lambda_C]^\top$ with $\lambda_c > 0$ ($c = 1, \ldots, C$) is the prior parameter. The instance-level class indicator $\mathbf{z}$ is a binary $C$-vector with the 1-of-$C$ code ($z_c = 1$ if the $c$-th class is chosen; $\forall i \neq c$, $z_i = 0$), which is generated from a multinomial distribution with parameter $\tilde{\boldsymbol{\theta}} = \boldsymbol{\theta}/\|\boldsymbol{\theta}\|_1$. The example-level label $\mathbf{y} = [y_1, \ldots, y_C]^\top$ is also a binary $C$-vector with $y_c = 1$ if the pattern $\mathsf{X}$ belongs to the $c$-th class (i.e., $c \in \mathsf{Y}$) and $y_c = 0$ otherwise. We assume the example-level label is generated by a cost-sensitive voting process based on the instance-level labels. Denote the average of the instance-level label assignments $\bar{\mathbf{z}} = [\bar{z}_1, \ldots, \bar{z}_C]^\top$, where $\bar{z}_c = \frac{1}{M}\sum_{m=1}^M z_{mc}$, we use a degraded soft-max regression model:

$$p(y_c|\bar{\mathbf{z}}, \mathbf{w}) \propto \exp(y_c w_c \bar{z}_c). \quad (1)$$

We assume that each instance $\mathbf{x}$ is described by a *bag of discrete features* $\{f^{(1)}, \ldots, f^{(K)}\}$, although it is quite straightforward to substitute with other instance models, e.g. mixtures of Gaussian. Bag-of-feature representation for images has been popularized in computer vision due to its simplicity, robustness and effectiveness (Nowak et al, 2006). For AIA, this can be achieved by building a vocabulary of visual codewords, e.g., by clustering representative image patches (see Section 4 for details). Let $\mathcal{F} = \{f_1, f_2, \ldots, f_D\}$ be the dictionary of discrete features, we therefore have a multinomial model:

$$p(\mathbf{x}|\mathbf{z}, B) \propto \beta_{c1}^{x_1} \beta_{c2}^{x_2} \ldots \beta_{cD}^{x_D}|_{z_c=1}, \quad (2)$$

where $x_d$ ($d = 1, \ldots, D$) is the frequency of the $d$-th feature $f_d$ in $\mathbf{x}$, $K$ is the total frequency of all features: $K = \sum_{d=1}^D x_d$, and $B = [\boldsymbol{\beta}_1, \ldots, \boldsymbol{\beta}_C]^\top$ is a $C \times D$-matrix with the $(c, d)$-th entry $\beta_{cd} = p(f_d = 1|z_c = 1)$.

The joint distribution of an example $(\mathsf{X}, \mathsf{Y})$ and the latent variables $\boldsymbol{\theta}$ and $\{\mathbf{z}\}$, given the model parameters $B, \boldsymbol{\lambda}$ and

**w**, is defined by:

$$p(\mathsf{X}, \mathsf{Y}, \{\mathbf{z}\}, \boldsymbol{\theta}|B, \boldsymbol{\lambda}, \mathbf{w}) \qquad (3)$$
$$= p(\boldsymbol{\theta}|\boldsymbol{\lambda}) \prod_{m=1}^{M} \left( p(\mathbf{z}_m|\boldsymbol{\theta}) \prod_{d=1}^{D} p(f_{md}|B, \mathbf{z}_m) \right) p(\mathbf{y}|\bar{\mathbf{z}}, \mathbf{w})).$$

As the number of classes, $C$, is typically very large in M$^3$C, the parameters, $\boldsymbol{\lambda}$ and $B$, could have very high dimensions. This usually leads to severe problems of overfitting when maximum likelihood estimators are used. To cope with this problem, we adopt a fully Bayesian treatment by posing conjugate priors over these variables. Particularly, we assume that $\{\lambda_1, \ldots, \lambda_C\}$ are *i.i.d.* generated from a Gamma distribution: $\lambda_c \sim \text{Gamma}(\boldsymbol{\chi})$, $\boldsymbol{\chi} = [\chi_1, \chi_2]^\top$; and $\{\boldsymbol{\beta}_1, \ldots \boldsymbol{\beta}_C\}$ are sampled from a Dirichlet distribution $\boldsymbol{\beta}_c \sim \text{Dir}(\boldsymbol{\eta})$, where $\boldsymbol{\eta} = [\eta_1, \ldots, \eta_D]^\top$ with $\eta_d \geqslant 0$, $c = 1, \ldots, C, d = 1, \ldots, D$. The overall model is depicted with a graphical representation in Figure 1.

### 3.3 Max-Margin Learning

Traditional topical mixture models are usually trained by MLE style estimators (e.g., variational EM, expectation propagation) or Gibbs sampling, where the likelihood of the observations are approximately maximized (Blei et al, 2003; Blei and Mcauliffe, 2008):

$$\max_{\boldsymbol{\lambda}, \boldsymbol{\eta}, \mathbf{w}} \quad L = \log p(\mathcal{D}|\boldsymbol{\lambda}, \boldsymbol{\eta}, \mathbf{w})$$

In this paper, by casting topics explicitly as class labels, we are eligible for discriminative learning. Here we develop maximum margin algorithms for inference and learning of EMM, leading to a hybrid generative/discriminative approach. By taking advantage of max-margin learning's well-known ability in handling small-sample problem, we hope to relieve the overfitting issue incurred by the massive classes in M$^3$C. We also hope to improve the classification accuracy by directly maximizing the discriminative ability using max-margin optimization. None of these concerns could be otherwise addressed by MLE learning.

The formulation is based on margin-maximization, a principle used in training SVMs. Our algorithm is an integration of *variational inference* (Jaakkola and Jordan, 2000) and *pariwise ordinal regression* (Herbrich et al, 1999), attempting to find parameters that place decision boundaries between each *relevant vs. irrelevant* label pair as far apart as possible. The learning problem is formulated as follows:

$$\min_{\boldsymbol{\lambda}, \boldsymbol{\eta}, \mathbf{w}} \left( \frac{\nu_1}{2} \mathbf{w}^\top \mathbf{w} - L + \frac{\nu_2}{N} \sum_n \frac{1}{|\mathsf{Y}_n||\mathsf{Y}_n^\circ|} \sum_{i,j} \xi_n^{i,j} \right) \qquad (4)$$
$$s.t.: \mathbb{E}[w_i \bar{z}_{ni} - w_j \bar{z}_{nj}] \geqslant 1 - \xi_n^{ij}, \forall i \in \mathsf{Y}_n, j \in \mathsf{Y}_n^\circ$$
$$\nu_1 \geqslant 0, \ \nu_2 \geqslant 0, \ \xi_n^{ij} \geqslant 0,$$

where $\nu_1$ and $\nu_2$ are trade-off parameters, the irrelevant label set is the complement of the relevant label set: $\mathsf{Y}_n^\circ = \mathcal{Y} - \mathsf{Y}_n$, and $L$ denotes the log likelihood for observing the labeled samples $\mathcal{D} = \{\mathsf{X}_n, \mathsf{Y}_n\}_{n=1}^N$.

Our discriminative formulation enjoys two additional advantages: (1) it exploits stronger supervision – suppose a training example has $m$ relevant tags and $n$ irrelevant ones, the pairwise formulation extends supervision size from $\mathcal{O}(m+n)$ to $\mathcal{O}(m \times n)$; (2) it relieves the data imbalance issue – typically, in a massive (e.g. thousands) tag set, relevant tags per example only occupy a tiny proportion (e.g., 3-5 per example), thus the MLE formulation suffers severe supervision imbalance (i.e., the prior positive-negative ratio $m/n \ll 1$), whereas our max-margin formulation is perfectly balanced, i.e.: a priori, each pair is equally positive (correctly ranked) or negative (incorrectly ranked).

The Lagrangian of Eq(4) is given by:

$$\mathfrak{L} = \frac{\nu_1}{2} \mathbf{w}^\top \mathbf{w} + \frac{\nu_2}{N} \sum_n \frac{1}{|\mathsf{Y}_n||\mathsf{Y}_n^\circ|} \sum_{i,j} \xi_n^{i,j}$$
$$- L + \sum_{nij} \alpha_n^{ij} (1 - \xi_n^{ij} - \mathbb{E}[w_i \bar{z}_{ni} - w_j \bar{z}_{nj}]),$$

where $\alpha_n^{ij} \geqslant 0$ are Lagragian multipliers. As both the last two terms (the likelihood objective $L$ and the constraints) involve marginal probabilities that requires integration over the latent variables, the exact computation of which is not tractable, we therefore use mean-field variational method to derive an upper bound to approximate the Lagrangian objective, and then optimize the upper bound instead. The overall learning algorithm is a EM optimization, where the E-step uses variational method to approximate the Lagrangian, whereas the M-step in turn optimizes it to learn the model parameters.

#### 3.3.1 Variational Approximation

We bound the intractable terms in the Lagrangian by applying the mean-field variational method (Jaakkola and Jordan, 2000). Particularly, for the loglikelihood $L$, we have:

$$L = \log p(\mathcal{D}|\boldsymbol{\lambda}, \boldsymbol{\eta}, \mathbf{w})$$
$$= \sum_n \log \int_{\boldsymbol{\theta}} \sum_{\{\mathbf{z}\}} p(\mathsf{X}_n, \mathsf{Y}_n, \{\mathbf{z}\}_n, \boldsymbol{\theta}_n) d\boldsymbol{\theta}$$
$$= \mathcal{L}(\boldsymbol{\gamma}, \Phi, \boldsymbol{\mu}, \boldsymbol{\rho}) + KL(q||p)$$
$$\approx \max_{\boldsymbol{\gamma}, \Phi, \boldsymbol{\mu}, \boldsymbol{\rho}} \mathcal{L}(\boldsymbol{\gamma}, \Phi, \boldsymbol{\mu}, \boldsymbol{\rho}), \qquad (5)$$

where we have introduced a full-factorized variational distribution: $q = \prod_c \text{Dir}(\boldsymbol{\beta}_c|\boldsymbol{\mu}_c) \prod_n (\text{Gamma}(\boldsymbol{\theta}_n|\boldsymbol{\gamma}_n, \boldsymbol{\rho}_n) \prod_m \text{Mult}(\mathbf{z}_{nm}|\boldsymbol{\phi}_{nm}))$, $KL(q||p)$ is the Kullback-Leibler (KL) divergence between $q$ and the posterior distribution of the latent variables. Denote $\mathcal{H}_q$ the entropy of $q$, $\mathcal{L}$ is the variational lower bound for $L$:

$$\mathcal{L}(\boldsymbol{\gamma}, \Phi, \boldsymbol{\mu}, \boldsymbol{\rho}) = \mathbb{E}_q[\log p(\mathcal{D}, \{\mathbf{z}\}, \{\boldsymbol{\theta}\}, \{\boldsymbol{\beta}\}|\boldsymbol{\lambda}, \mathbf{w}, \boldsymbol{\eta})] + \mathcal{H}_q.$$

The main terms in the variational bound $\mathcal{L}$ are given by:

$$\mathbb{E}_q[\log p(\boldsymbol{\theta}|\boldsymbol{\lambda})] = \sum_{c=1}^{C}(\log \lambda_c - \lambda_c \gamma_c \rho_c)$$

$$\mathbb{E}_q[\log p(\mathbf{z}_m|\boldsymbol{\theta})] = \sum_{c=1}^{C} \phi_{mc}(\Psi(\gamma_c) + \log \rho_c)$$

$$\mathbb{E}_q[\log p(\mathsf{X}|B, \{\mathbf{z}\})] = \sum_{m=1}^{M}\sum_{c=1}^{C}\sum_{d=1}^{D} \phi_{mc} x_{md} \log \beta_{cd}$$

$$\mathbb{E}_q[\log p(\mathbf{y}|\bar{\mathbf{z}}, \mathbf{w})] = \frac{1}{M}\sum_{m=1}^{M}\sum_{c=1}^{C} y_c w_c \phi_{mc}$$

$$\mathcal{H}_q[\boldsymbol{\theta}] = \sum_{c=1}^{C}(\log \rho_c + \gamma_c + \log \Gamma(\gamma_c) - (\gamma_c - 1)\Psi(\gamma_c))$$

where $\Psi(\cdot)$ is the digamma function, and we have used the fact $\mathbb{E}_q[\log \theta_c] = \Psi(\gamma_c) + \log \rho_c$. The other terms are standard in the topical model literature (Blei et al, 2003).

The Lagrangian $\mathfrak{L}$ is therefore approximated by solving:

$$\max_{\boldsymbol{\gamma},\boldsymbol{\Phi},\boldsymbol{\mu},\boldsymbol{\rho}} \quad \mathcal{L} + \sum_{nij} \alpha_n^{ij}(w_i \mathbb{E}_q[\bar{z}_{ni}] - w_j \mathbb{E}_q[\bar{z}_{nj}]). \quad (6)$$

where $\mathbb{E}_q[\bar{z}_{ni}] = \frac{1}{M_n}\sum_{m=1}^{M_n} \phi_{nmi}$. We use a coordinate ascent method, which leads to the following iterative updating equations:

$$\rho_{nc} = \frac{\lambda_c \gamma_{nc}}{1 + \sum_{m=1}^{M_n} \phi_{nmc}} \quad (7)$$

$$\phi_{nmc} \propto \rho_{nc} \prod_{d=1}^{D}(\beta_{cd})^{x_{nmd}} e^{\Psi(\gamma_{nc}) + \frac{w_c}{M_n}(y_{nc} + \delta_{nc})} \quad (8)$$

$$\mu_{cd} = \eta_d + \sum_{n=1}^{N}\sum_{m=1}^{M_n} \phi_{nmc} x_{nmd} \quad (9)$$

where $\delta_{nc}$ is a quantity related to the Lagrangian multipliers $\alpha$'s due to the margin constraints, we have:

$$\delta_{nc} = \begin{cases} \sum_{j \notin \mathsf{Y}_n} \alpha_n^{cj}, & \text{if } c \in \mathsf{Y}_n \\ -\sum_{i \in \mathsf{Y}_n} \alpha_n^{ic}, & \text{else} \end{cases}$$

Note that there is no closed-form update for $\gamma_{nc}$, instead, they are solved using Newton-Raphson algorithm from a nonlinear equation:

$$(\sum_{m} \phi_{nmc} - \gamma_{nc} + 1)\Psi'(\gamma_{nc}) + 1 - \lambda_c \rho_{nc} = 0. \quad (10)$$

It is worth mentioning that by using exponential (instead of Dirichlet) topic mixtures, our model is able to identify relevant labels more effectively. This could be seen from Eq.(4), in which:

$$-L_{\theta|\lambda} \propto ||\boldsymbol{\lambda} \circ \boldsymbol{\theta}||_1 = \sum_{c=1}^{C} \lambda_c \theta_c.$$

As in *LASSO* (Tibshirani, 1994) and *sparse-coding* (Lee et al, 2009), this weighted $\ell_1$ regularization will naturally lead to sparse topic mixtures $\boldsymbol{\theta}$, that is, although the dimension of $\boldsymbol{\theta}$ (i.e. the total number of classes $C$) might be huge, only a few of its components $\theta_c$ have nonzero values[1], therefore, it effectively reduces the number of candidate classes to which an given example $\mathsf{X}$ is relevant. Considering that the instance-level labels $\mathbf{z}$'s are sampled according to $\boldsymbol{\theta}$ and the example-level label $\mathbf{y}$ is generated based on $\bar{\mathbf{z}}$, the exponential membership is acting like a filter which automatically identifies a few relevant labels while ruling out a large number of irrelevant ones.

### 3.3.2 Parameter Estimation

The optimization in Eq.(4) is decomposable to the three parameters $\boldsymbol{\lambda}, \boldsymbol{\eta}$ and $\mathbf{w}$. Since $\boldsymbol{\lambda}$ and $\boldsymbol{\eta}$ [2] are only involved in the likelihood $L$, we can learn these parameters by an approximate empirical Bayesian procedure, i.e., maximizing the variational likelihood $\mathcal{L}$ while keeping the variational parameters fixed. Specifically, taking into account the Gamma prior over $\lambda$, we have:

$$\lambda_c = (\chi_1 + N - 1)/(\chi_2 + \sum_{n=1}^{N} \gamma_{nc}\rho_{nc}),$$

whereas the Dirichlet smooth prior $\eta$ can be learned again by a Newton-Raphson procedure, similar to what is used in the standard LDA model (Blei et al, 2003).

To learn $\mathbf{w}$, we need to solve the quadratic programming (QP) Eq.(4), where

$$\mathbb{E}[w_i \bar{z}_{ni}] = \frac{1}{M_n} w_i \sum_{m=1}^{M_n} \phi_{nmi}.$$

In this paper, we use $SVM^{perf}$ (Joachims, 2006), a cutting-plane SVM solver. Note that although Eq.(4) involves a huge number (order $\mathcal{O}(C^2 N)$) of constraints, only a little proportion of them are actually active (i.e. support vectors) because most of the Lagrangian multipliers $\alpha$'s will become exact zero at optima. Also, by cutting-plane optimization, which progressively adds only the most violated constraints, we can guarantee that the working set is always of a controllable size.

### 3.3.3 Prediction

We consider both task (1) and (2) mentioned in §1. For task (1), i.e., label (caption) prediction for a given example (image) $\mathsf{X}$, we first run the variational inference procedure, and approximate the label distribution by:

$$p(\mathsf{Y}|\mathsf{X}) \approx \hat{\boldsymbol{\theta}}^{\mathbf{w}}/||\hat{\boldsymbol{\theta}}^{\mathbf{w}}||_1.$$

where $\hat{\theta}_c^w = w_c \hat{\theta}_c$, and $\hat{\theta}_c$ is inferred by an MAP estimator:

$$\hat{\theta}_c = [(\gamma_c - 1)\rho_c]^+.$$

---

[1] If we adopt a *max a posterior* (MAP) inference for $\theta$, from Eq(10), non-zero $\theta_c$ can only be achieved by class $c$ satisfying $\sum_m \phi_{mc} > \frac{\lambda_c \rho_c - 1}{\Psi'(\gamma_c)}$.

[2] We implemented exchangeable Dirichlet, i.e., $\forall d$: $\eta_d = \eta$.

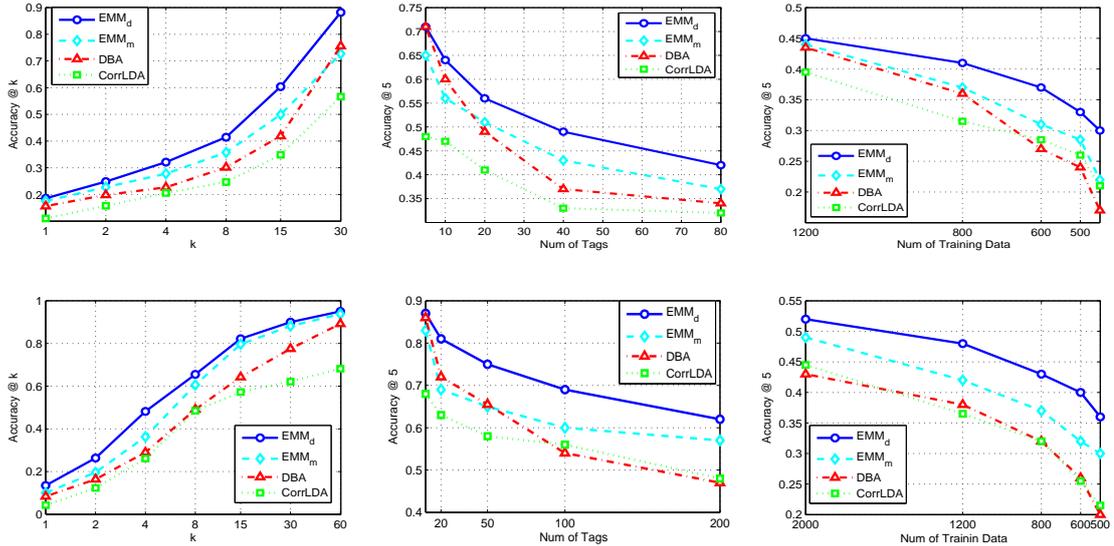

Figure 2: Image annotation performance and tag-scalability comparison. (Left) Top-$k$ accuracy vs. $k$. (Middle) Top-5 Accuracy vs. tag size. (Right) Top-5 Accuracy vs. training set size. (Top) Results on Alipr. (Bottom) Results on LabelMe.

Note that all the irrelevant tags $\{c : \sum_m \phi_{mc} \leqslant \frac{\lambda_c \rho_c - 1}{\Psi'(\gamma_c)}\}$ would lead to exact zero response $\theta_c = 0$. Also note that, the exponential term $\exp(\frac{w_c}{M_n}(y_{nc} + \delta_{nc}))$ should be removed when updating $\phi$ in Eq.(8) since the label of a testing example X is unobserved.

For the task (2), i.e., label (tag) prediction for each instance (image region), we have:

$$p(\mathbf{z}_m|\cdot) = \int_{\boldsymbol{\theta}} p(\mathbf{z}_m, \boldsymbol{\theta}|\cdot) d\boldsymbol{\theta} \approx q(\mathbf{z}_m|\boldsymbol{\phi}_m)$$

That is , we first run variational inference on X (unlabeled image) or (X, Y) (labeled image), then predict the instance label by $p(z_{mc} = 1) = \phi_{mc}$.

## 4 Experimental Results

We evaluate our method on two data sets that are from popular online annotation engines Alipr (Li and Wang, 2008) and LabelMe (Russell et al, 2007) respectively. The Alipr data consists of 2359 images and a total number of 142 unique tags, on average there are 2.02 tags per image. The LabelMe data set contains 4053 images and 1017 tags, 7.81 tags per image. Both of them cover sufficiently diverse scenes, e.g., indoor, urban, village, road and landscape.

For each image-caption pair, the image is segmented by the N-Cuts algorithm (Shi and Malik, 2000), each region is used as an instance and each word in the caption is used as a label[3]. The geometry-free bag-of-discrete-feature model is then adopted to represent each region due to its simplicity, robustness and good performance (Nowak et al, 2006).

---
[3]We only keep tags that present more than three times, leaving with us 93 tags on Alipr and 291 tags on LabelMe.

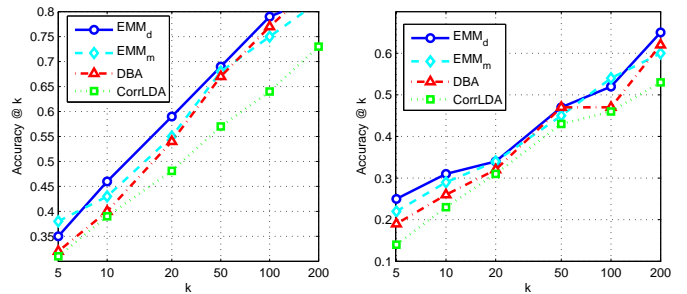

Figure 3: Region annotation performance comparison: Top-$k$ accuracy vs. $k$ for annotated regions in captioned images (left) and images without captions (right). Both results are obtained on LabelMe data set.

Particularly, we randomly sample a fixed-size (1000) set of local patches (scale of each patch is chosen randomly between 5×5 to 1/4 of the image size) from each image, characterize each patch using the 128-dimensional SIFT descriptor (8 orientations, 4×4 blocks of 3×3-scaled cells) (Lowe, 2004), and encode the descriptor using hard membership to the nearest codebook center. In this way, each image is analogous to a fixed-length (1000 words) document, each region to a paragraph and represented by a histogram of visual word counts. The visual vocabulary is built by running the $k$-mean algorithm ($k = 1000$; 5 repetitions with random initialization) to vector quantize descriptors of patches sampled from randomly chosen 500 images.

We tested both discriminatively trained EMM (EMM$_d$) and MLE trained EMM (EMM$_m$). For comparison, we considered only models that are capable to model ambiguity. Particularly, we compared with two approaches: one is the

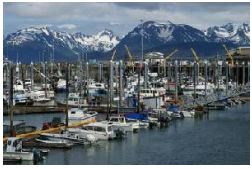 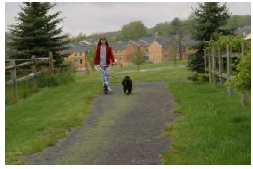 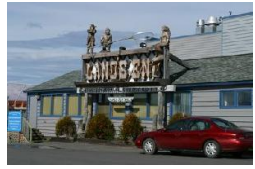 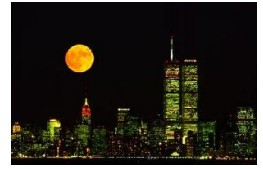

snow mountain, cloud, boat, harbor | house, lawn, people, tree | house, car | building, night
water, boat, snow, mountain, sky | people, tree, house, road, bear | car, water, people, tree, cloud | building, sun, flower, fireworks, night
snow, water, mountain, sky, plane | tree, house, road, people, outdoor | car, tree, house, water, sky | building, sun, flower, sky, tree
mountain, plane, house, water, indoor | tree, people, bridge, plant, garden | house, car, bird, water, urban | sun, building, cloud, pattern, water
water, mountain, snow, plant, bridge | people, tree, cloud, sky, road | car, water, house, bear, sign | building, land, sun, tree, animal

Figure 4: Example annotations on Alipr data by different methods: (from top to bottom) true caption, the hybrid approach (EMM$_d$), EMM$_m$, CorrLDA and DBA.

CorrLDA model (Blei and Jordan, 2003) (200 topics were used), a machine-translation-based method that have been shown to outperform many methods of this class (Blei and Jordan, 2003; Wang et al, 2009); the other is DBA (Yang et al, 2010), a MLE-trained Dirichlet-Multinomial mixture model for multi-label multi-instance classification. The trade-off parameters $\nu_1$ and $\nu_2$ are tuned by a 5-fold cross validation procedure. The reported results of testing accuracy are also estimated by 5-fold cross validation.

We first apply all these approaches to image-level annotation, i.e., image caption prediction. As all the models output probabilistic predictions, we use the Top-$k$ accuracy as evaluation metric, which is the Micro-averaged $F_1$ measure computed on the top-ranked $k$-sublist of the predictions. The results are illustrated in Figure 2 (the leftmost column). Firstly, we observe that classification based approaches generally achieves better performance than machine-translation based approach, validating that it is beneficial solving AIA as a classification task. Among the three classification approaches, the hybrid approach (the discriminatively trained EMM) achieves consistently the best accuracies $\forall k = 1, \ldots, 40$ on both data sets. Interestingly, between the two MLE trained classifiers, on Alipr, EMM$_m$ performs better than DBA for the head (i.e., $k < 19$) predictions and slightly worse for tail predictions (larger $k$), whereas on LabelMe, whose tag size is bigger, EMM$_m$ performs almost consistently for the top 40 predictions, indicating that the EMM model might have better tag-scalability. It is worth noting that both the two measures (i.e., prediction sparsity and max-margin learning) turns out to significantly improve the performances, particularly, in terms of Accuracy@5: (1) by encouraging prediction sparsity, the EMM model alone (EMM$_m$ vs DBA) improves the performance by 18.84% on Alipr and 22.79% on LabelMe; (2) by max-margin learning, the hybrid approach (EMM$_d$) further gains 9.08% (on Alipr) or 15.04% (on LabelMe) improvements over EMM$_m$. The annotation results of our AIA system is satisfactorily good. As an intuitive demonstration, Figure 4 shows some example images and the top-5 annotations by each algorithm. As we can see, most of the annotations by the hybrid approach are reasonably relevant to the image scenes, much better than other competitors. It even accurately identified the global theme of a scene (e.g., "night" in the rightmost graph).

We further test the label-scalability of each algorithm. Also shown in Figure 2 is the Top-5 accuracy of each algorithm trained on (1) fixed-size image set with increasing-number of tags; and (2) decreasing-size image set with a fixed number of tags. As expected, the performances of DBA deteriorate exponentially for both cases. In contrast, although the hybrid approach is also based on classification, its performance is very stable. In fact, the scalability of EMM$_d$ is much better even compared to the machine translation based approach CorrLDA.

We finally compare the three algorithms on region-level annotation. We conduct this task for both (1) captioned images and (2) uncaptioned images. Note that the Alipr data set does not include region-level labels (ground truth) and hence cannot be used to assess the performances in this task. The results on LabelMe data are depicted in Figure 3. For the first setting, the hybrid approach outperforms the other competitors consistently; for the latter, it performs the best for the top-relevant predictions and comparably for tail predictions, which is still quite encouraging because the top predictions are usually the most important, upon which the final tags are finally decided. We also observe that all the classification-based algorithms, EMM$_d$, EMM$_m$ and DBA, substantially outperform the machine-translation-based approach CorrLDA, suggesting that the former are more powerful for disambiguating scenes.

## 5 Conclusion

Automatic image annotation involves modeling data that are both extremely ambiguous (both input and output) and sparse (e.g., too many tags making data too scarce for learning each), critically challenging existing learning algorithms. In this paper, we investigated the generic task of "Many-Class Multi-Label Multi-Instance Classification" (M$^3$C) and devised a hybrid generative-discriminative learning approach to M$^3$C tasks such as AIA. The proposed approach includes a Bayesian Hierarchical model, which is able to capture both the input and output ambiguity and in the meanwhile encourage prediction sparseness, and a discriminative learning formulation, which integrates the variational inference and pairwise ordinal regression to maximize the prediction power. We tested our approach on two real-world benchmarks and showed satisfactory annotation performance as well as superior scalability to the tag size.

One limitation of the current model is that, by assuming instance exchangeability, it does not account for context correlations, for example, the tag "Apple" is more likely to mean *computer* rather than *fruit* if it appears in a scene together with "CDs", "USB" and "mouse". We plan to explore such context correlation in future work. Also, the scale of our evaluation is limited by the availability of labeled data, we plan to extend it to real-world scale (e.g., tens of thousand of tags) as soon as larger labeled data are available. We also plan to empirically compare our model with state-of-art annotation algorithms (e.g., weighted $k$NN, (Verbeek et al, 2010)) to provide a big picture of existing annotation methods.

**Acknowledgements**

We thank the anonymous reviewers for helpful comments. Part of this work is supported by NSF #DMS-0736328, NSFC #60828001, the 111-Project #B07022 and a grant from Microsoft.